%% file: main.tex
\documentclass{article}
\usepackage[table]{xcolor}
\usepackage[preprint]{corl_2025} 

\usepackage{pifont}
\usepackage{xspace}
\usepackage{booktabs}
\usepackage{adjustbox}
\usepackage{caption}
\usepackage{multirow}
\usepackage{float}
\usepackage{graphicx}
\usepackage{balance}
\usepackage{paralist}
\usepackage{siunitx}
\usepackage{wrapfig}
\usepackage{enumitem}

\newcommand{\xmark}{\color{deepred}{\ding{55}}}
\newcommand{\cmark}{\color{deepgreen}{\ding{51}}}

\newcommand{\ours}[0]{GMT\xspace}

\input{math_commands}

\definecolor{deepgreen}{RGB}{63, 126, 49}
\definecolor{deepred}{RGB}{196, 49, 25}
\definecolor{deepblue}{RGB}{0, 0, 139}

\title{\LARGE{GMT: {\color{deepred}G}eneral {\color{deepred}M}otion {\color{deepred}T}racking for \\ Humanoid Whole-Body Control}}

\author{
  Zixuan Chen$*^{1,2}$ \And Mazeyu Ji$*^{1}$ \And Xuxin Cheng$^{1}$ \And Xuanbin Peng$^{1}$ \AND
  Xue Bin Peng\dag$^{2}$ \And Xiaolong Wang\dag$^{1}$ \AND
  $^{1}${\normalfont UC San Diego} \And $^{2}${\normalfont Simon Fraser University} \And $*${\normalfont Equal Contribution} \And {\normalfont \dag Equal Advising}
  \vspace{0.3cm}
  \AND \href{https://gmt-humanoid.github.io}{\textcolor{deepred}{\large gmt-humanoid.github.io}}
  }

\begin{document}
\maketitle

\vspace{-1cm}
\begin{figure}[ht]
        \centering
        \includegraphics[width=1.0\textwidth]{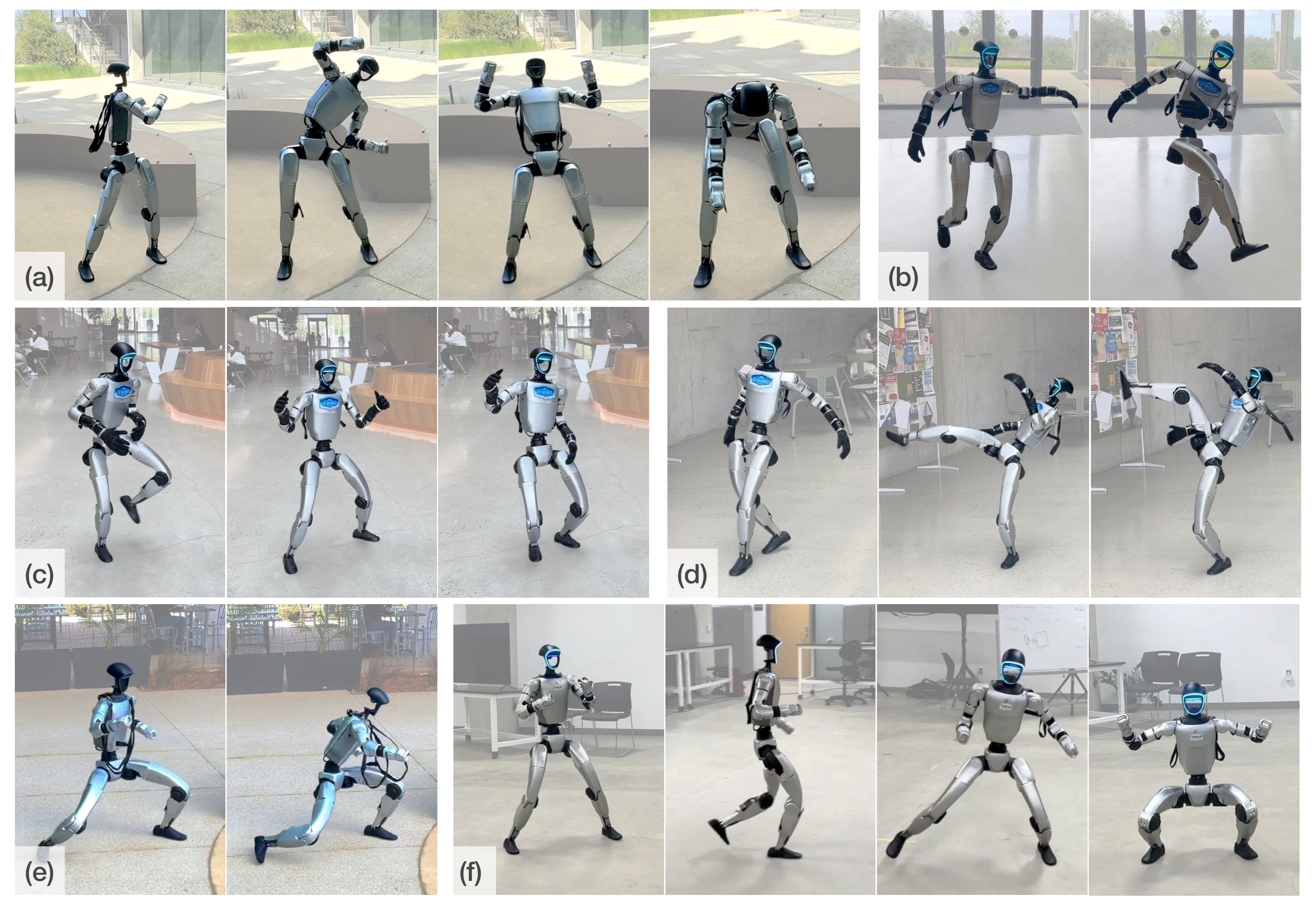}
        \caption{We deploy the \textit{general} unified motion tracking policy on a medium-sized humanoid robot. 
        \ours can perform a wide range of motion skills with good stability and generalizability, including \textbf{(a)} stretching, \textbf{(b)} kicking-ball, \textbf{(c)} dancing, \textbf{(d)} high kicking, \textbf{(e)} kungfu, and \textbf{(f)} other dynamic skills such as boxing, running, side stepping, and squatting.
        }
        \label{fig:teaser}
\vspace{-0.25cm}
\end{figure}

\begin{abstract}
    The ability to track general whole-body motions in the real world is a useful way to build general-purpose humanoid robots.
    However, achieving this can be challenging due to the temporal and kinematic diversity of the motions, the policy's capability, and the difficulty of coordination of the upper and lower bodies. 
    To address these issues, we propose \textbf{\ours}, a general and scalable motion-tracking framework that trains a single unified policy to enable humanoid robots to track diverse motions in the real world. \ours is built upon two core components: an Adaptive Sampling strategy and a Motion Mixture-of-Experts (MoE) architecture. 
    The Adaptive Sampling automatically balances easy and difficult motions during training. The MoE ensures better specialization of different regions of the motion manifold.
    We show through extensive experiments in both simulation and the real world the effectiveness of \ours, achieving state-of-the-art performance across a broad spectrum of motions using a \textit{unified general} policy. Videos and additional information can be found at \href{https://gmt-humanoid.github.io}{\textcolor{deepred}{\texttt{gmt-humanoid.github.io}}}.
    
\end{abstract}



\section{Introduction}
    One of the primary goals for humanoid robots is to perform a wide range of tasks in everyday environments. Enabling humanoid robots to produce a broad repertoire of human-like movements is a promising approach towards achieving this objective.
    To generate such human-like movements, a general whole-body controller is required — one capable of leveraging a wide corpus of motor skills to perform both basic tasks, such as natural walking, and more dynamic and agile actions, such as kicking and running. With such a robust whole-body controller, it becomes possible to integrate a high-level planner that selects and sequences skills autonomously, paving the way towards general-purpose humanoid robots.

    Manually designing such controllers can be challenging and labor-intensive due to the high degrees of freedom (DoFs) in humanoid systems and the complexity of real-world dynamics. Since humanoid robots are designed to closely resemble the human body, human motion data offers an ideal resource for equipping robots with a rich set of skills. In the field of computer graphics, researchers have leveraged human motion data and learning-based methods to develop a single unified controller for simulated characters that can reproduce a wide variety of motions and perform diverse skills with remarkably human-like behaviors~\citep{peng2018deepmimic, peng2022ase, luo2023perpetual, truong2024pdp, tessler2024maskedmimic}.

\begin{wrapfigure}{r}{0.43\textwidth}
        \includegraphics[width=0.42\textwidth]{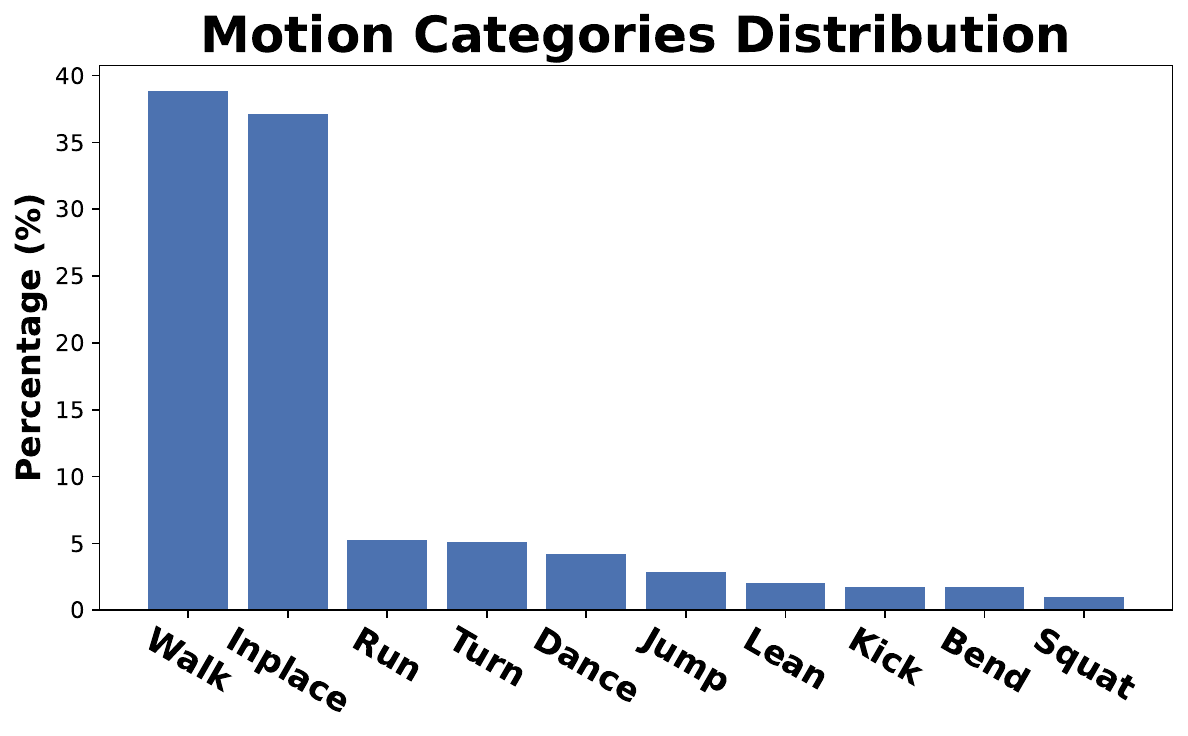}
        \caption{Distribution of motion categories in the AMASS dataset. The figure shows the proportion of the total motion duration corresponding to each category.}
        \label{fig:motion-distribution}
    \end{wrapfigure}

    Despite great progress in simulated domains, developing such a generic unified controller for humanoid robots 
    can be challenging due to:
    \begin{itemize} 
        \item \textbf{Partial Observability.} For real-world robots, full state information like linear velocities and global root positions is not accessible, which is quite important when learning motion tracking policies. The absence of this information makes the training process more challenging.
        \item \textbf{Hardware Limitations.} Existing human motion datasets often include movements such as back-flipping and rolling, which are infeasible for humanoid robots to execute due to hardware limitations. Moreover, even for more basic skills like walking and running, robots may struggle to produce enough torque to accurately match the speed and dynamics of human motion. These mismatches require extra special handling during the training of motion tracking controllers for robots.
        \item \textbf{Unbalanced Data Distribution.} Large mocap datasets like AMASS~\citep{AMASS:ICCV:2019} often exhibit highly unbalanced distributions, as shown in \Figref{fig:motion-distribution}, with a large proportion of motions involving walking or in-place activities. The scarcity of more complex or dynamic motions can cause the robot to struggle in mastering these less frequent but critical skills.
        \item \textbf{Model Expressiveness.} While a simple MLP-based network may be sufficient to develop a satisfying control policy when tracking a few motion clips, it often struggles when applied to large Mocap datasets. Such architectures typically lack the capacity to capture complex temporal dependencies and to distinguish between diverse motion categories. This limited expressiveness can result in sub-optimal tracking performance and poor generalization across a wide range of skills.
    \end{itemize}
    Although existing works have tried to solve some individual issues mentioned above, for example, teacher-student training framework is used to handle partial observability~\citep{ji2024exbody2, he2024omnih2o}; several categories of small datasets are used to fine-tune separate specialist policies~\citep{ji2024exbody2}; and a transformer model is used to increase the expressiveness of the model~\citep{fu2024humanplus}, developing a unified general motion tracking controller remains an open problem. This paper demonstrates that by addressing data distribution and model problems jointly, along with some other careful design decisions to handle partial observability and hardware issues, we can create an effective system for training general motion tracking controllers for real humanoid robots. A brief comparison of our work with highly related existing works is shown in \tabref{tab:work-comp}. We measure if a policy is \textit{diverse} based on its ability to perform a broad range of both upper-body and lower-body skills, including variations in walking styles, crouching, kicking, and other expressive whole-body movements.
    
    In this paper, we propose \ours, a general and effective framework that trains a single unified, high-quality motion tracking policy for real-world humanoid robots from large mocap datasets.
    Central to \ours are two key innovations: a novel \textit{Adaptive Sampling} strategy,
    designed to mitigate issues of struggling to learn less frequent motions arising from uneven motion category distributions, 
    and a motion \textit{Mixture-of-Experts (MoE)} architecture to enhance model expressiveness and generalizability. 
    \Figref{fig:teaser} showcases the real-world deployment of our policy, demonstrating its ability to reproduce a wide range of basic locomotion and agile skills with high fidelity, achieving state-of-the-art performance using a \textit{single} unified policy. For more details, please refer to our \href{https://gmt-humanoid.github.io}{project website}.

    \input{tables/work_comp}


\section{Related Works}
\label{sec:related-work}
    Developing a general whole-body controller that enables humanoid robots to perform a wide range of skills has long been a fundamental yet challenging problem, primarily due to the systems' high dimensionality and inherent instability. Traditional model-based methods 
    have achieved robust whole-body locomotion controllers for bipeds and humanoids with gait planning and dynamics modeling~\citep{miura1984dynamic, sreenath2011compliant, geyer2003positive}. However, designing such controllers can be labor-intensive and requires careful handling of complex dynamics. In recent years, learning-based approaches have made significant progress in building whole-body controllers. They either develop the controller with careful hand-designed task rewards~\citep{radosavovic2024humanoid, radosavovic2024learning, radosavovic2024real, gu2024advancing, chen2024lcp, ben2025homie, zhang2024adam, huang2025moe}, or with human motions as reference~\citep{he2025asap, he2024omnih2o, fu2024humanplus, ji2024exbody2, cheng2024expressive, he2024hover, he2024h2o}.

\subsection{Learning-based Humanoid Whole-Body Control}
This work focuses on whole-body control for humanoid robots. Previous studies have demonstrated that whole-body control policies, trained with manually designed rewards, can enable humanoid robots to perform locomotion skills such as walking~\citep{radosavovic2024humanoid, radosavovic2024learning, radosavovic2024real, gu2024advancing, chen2024lcp}, jumping~\citep{zhang2024wococo, zhuang2024humanoidparkour, xue2025hugwbc}, and fall-recovery behaviors~\citep{he2025getup, huang2025getup}.
However, these controllers are typically task-specific, requiring training separate policies with customized reward functions for each task. For instance, policies developed for walking are not easily transferable to tasks such as jumping or manipulation. 
In contrast, human motion data offers a promising way to develop general purpose controllers without having to design task-specific reward functions for each skill we want the agent to perform.

\subsection{Humanoid Motion Imitation}
Leveraging human motion data to develop human-like whole-body behaviors for robots has been extensively studied in character animation. Previous works have achieved high-quality and general motion tracking on simulated characters~\citep{peng2018deepmimic, luo2023perpetual, tessler2024maskedmimic, truong2024pdp}, and diverse motion skills for various downstream tasks~\citep{peng2021amp, peng2022ase, tessler2023calm, hassan2023synthesizing, yuan2023tennis}. However, due to the partial observability in the real world, developing such whole-body controllers for real robots~\citep{cheng2024expressive, lu2024mobile, fu2024humanplus, he2024omnih2o, ji2024exbody2, he2025asap, serifi2024vmp, mao2024uhc, liu2024opt2skill} can be challenging. For developing a unified general whole-body motion tracking controller, some works decouple upper-body and lower-body control to compromise between expressiveness and balance~\citep{cheng2024expressive, lu2024mobile}. HumanPlus~\citep{fu2024humanplus} and OmniH2O~\citep{he2024omnih2o} then successfully enabled whole-body motion imitation on a full-sized robot, but with unnatural movements in the lower body. ExBody2~\citep{ji2024exbody2} achieved better whole-body tracking performance but with several separate specialist policies. VMP~\citep{serifi2024vmp} demonstrated high-fidelity reproduction of a wide range of skills on the real robot, but its dependency on the mocap system during deployment limits its applicability in the wild.

\section{Learning General Motion Tracking Controllers}
\label{sec:methods}
\begin{figure}[t]
        \centering
        \includegraphics[width=0.98\textwidth]{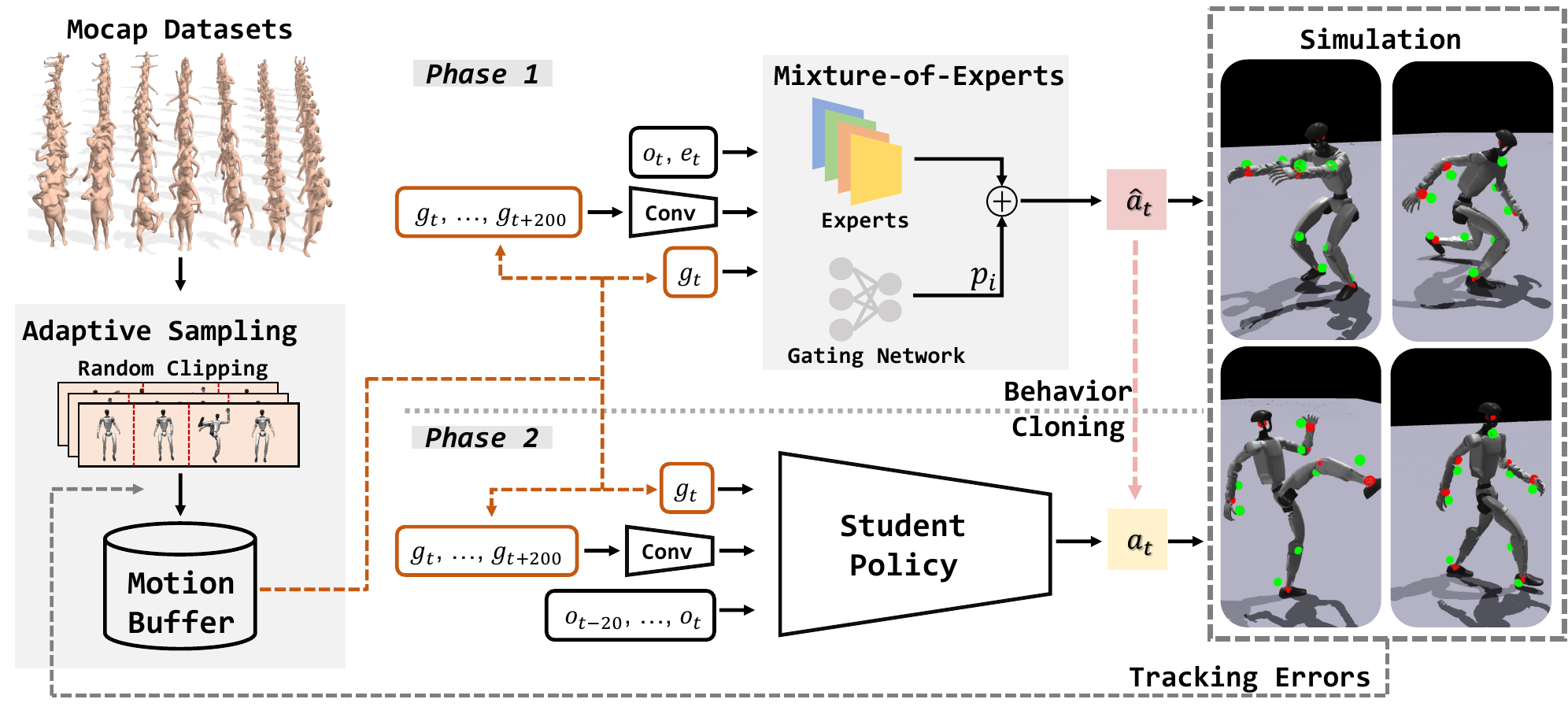}
        \caption{An overview of \ours. Here $\vg_t$ denotes the motion target frame, $\vo_t$ denotes proprioceptive observation, and $\ve_t$ denotes privileged information.}
        \label{fig:method-overview}
\end{figure}

In this section, we introduce \ours, a framework for developing general and high-quality motion tracking controllers for real-world humanoid robots. \ours follows the two-stage teacher-student training framework similar to prior works~\citep{ji2024exbody2, he2024omnih2o}, where a privileged teacher policy is first trained using PPO~\citep{schulman2017proximal}, and a student policy is then trained using DAgger\citep{ross2011dagger} by imitating the output of the teacher policy.
For the following part, we first introduce the core components of our method, which are \textit{Adaptive Sampling strategy} and \textit{Motion MoE architecture}. Then we present some key designs that are also crucial for policy learning, including motion input designs and the dataset curation process.

\subsection{Adaptive Sampling}
As illustrated in \Figref{fig:motion-distribution}, large motion datasets like AMASS~\citep{AMASS:ICCV:2019} exhibit significant category unbalance. This unbalance can substantially hinder the learning of less frequent and more complex motions. Moreover, longer sequences within such datasets are often composite motions, consisting of a series of skills that include both basic and challenging segments. Using previous sampling strategies~\citep{tessler2024maskedmimic, luo2023perpetual}, when a policy fails on harder segments, it is still trained on the entire motion sequence — often dominated by easier parts — which results in a low effective sampling rate for the difficult portions. 

To address these issues, we introduce \textit{Adaptive Sampling}, which consists of two key components:
\begin{itemize}
    \item \textbf{Random Clipping:}  Motions longer than $10$ seconds are clipped into several sub-clips, each with a maximum length of 10 seconds. To prevent artifacts at clip transitions, we apply randomized clipping by introducing a random offset of up to $2$ seconds. Additionally, all motions are re-clipped periodically during training to further diversify the sampled sub-clips;
    \item \textbf{Tracking Performance-based Probabilities:} During training, we log the completion level $c_{i}$ of each motion, and terminate episodes when the tracking error exceeds $E_{i}$. $c_{i}$ starts at $10$ and decays by multiplying $0.99$ each time the motion is completed, the minimum value is $1$. And $E_{i}=0.25\exp((c_{i}-1)/(10-1)*\log(0.6/0.25))$. We then define the sampling level $s_{i}$ of each motion as:
    \begin{equation}
    \label{eq: adaptive-sampling}
        \begin{aligned}
        s_i &=
        \begin{cases}
        \left( \min\left( E_{\text{max\_key\_body\_error}}/{0.15}, 1 \right) \right)^5, & c_i = 1, \\
        c_i, & c_i > 1.
        \end{cases}
        \end{aligned}
    \end{equation}
    The actual sampling probability is obtained by normalizing $s_{i}$.
\end{itemize}
 By applying Adaptive Sampling from the start of training, we avoid repeatedly sampling easy segments of long motions and focus training efforts on refining performance on harder motions with higher tracking errors.

\subsection{Motion Mixture-of-Experts}
To enhance the expressiveness of the model, we incorporate a soft MoE module during the training of the teacher policy. An illustration of our model is shown in \Figref{fig:method-overview}. The policy network consists of a group of expert networks and a gating network. The expert networks take both of the robot state observation and motion targets as input, which output the final action $\va_t$. The gating network also takes the same input observations and outputs a probability distribution over all experts. The final action output is a combination of actions sampled from each expert's individual action distributions:
$\va = \sum_{i=1}^{n}p_{i}\va_{i},$
where $p_{i}$ is the probability of each expert output by the gating network and $\va_{i}$ is the output of each expert policy.

\subsection{Dataset Curation}
We use a combination of AMASS~\citep{AMASS:ICCV:2019} and LAFAN1~\citep{harvey2020lafan} to train the motion tracking policy. Since raw datasets contain infeasible motions like crawling, fallen states, and extremely dynamic ones due to hardware constraints. As infeasible motions represent noise and can hamper learning, we adopt a two-stage data curation process similar to previous works~\citep{tessler2024maskedmimic}.
 In the first stage, we apply rule-based filtering to eliminate infeasible motions — for example, motions where the root's roll or pitch angles exceed specified thresholds, or where the root height is abnormally high or low. In the second stage, we train a preliminary policy on previously filtered dataset 
with approximately $5$ billion samples. Based on the completion rates achieved by this policy, we further filter out failed motions. This results in a curated version of the training dataset - a subset of AMASS and LAFAN1 with $8925$ clips totaling $33.12$ hours.

\subsection{Motion Inputs}
The goal of motion tracking is to let the robot track specific targets in each motion frame.
We represent the motion tracking target of each frame as: $\vg_t=[\vq_t, \vv_{t}^{\text{base}}, \vr_{t}^{\text{base}}, \vp_{t}^{\text{key}}, h_{t}^{\text{root}}]$, where $\vq_{t}\in\R^{23}$ represents joint positions, $\vv_{t}\in\R^{6}$ denotes the base linear and angular velocities, $\vr_{t}\in\R^{2}$ represents the base roll and pitch angles, $h_{t}^{\text{root}}$ represents root height, and $\vp_{t}^{\text{key}}\in\R^{3\times\text{num keybody}}$ corresponds to the local key body positions. Unlike prior works that use global key body positions~\citep{he2024omnih2o, he2025asap}, we adopt local key body positions similar to ExBody2~\citep{ji2024exbody2}, with the further refinement of aligning the local key bodies relative to the robot's heading direction.

Furthermore, to improve tracking performance, we move beyond using only the immediate next motion frame as input~\citep{ji2024exbody2, cheng2024expressive, he2024omnih2o}. Instead, we stack multiple consecutive frames $[\vg_{t}, \dots, \vg_{t+100}]$ covering approximately two seconds of future motion. These stacked frames are then compressed by a convolutional~\citep{krizhevsky2012imagenet} encoder into a latent vector $\vz_{t}\in\R^{128}$, which is then combined with the immediate next frame $\vg_{t}$ and fed into the policy network. This design enables the policy to capture both the long-term trends of the motion sequence and explicitly recognize the immediate tracking target. We show in experiments that this design is essential to high-quality tracking.

\section{Experiments}
\label{sec:experiments}
\subsection{Experimental Setups}
We evaluate \ours in both simulation and real-world settings. For simulation experiments, each policy is trained using approximately $6.8$ billion samples with domain randomization~\citep{peng2018dynamicsrand} and action delay~\citep{rudin2022learning}, on a filtered subset of the AMASS and LAFAN1~\citep{harvey2020lafan} datasets; and evaluated on AMASS test set~\citep{luo2023perpetual} and LAFAN1. We use IsaacGym~\citep{makoviychuk2021isaacgym} as the simulator and the number of parallel environments is $4096$. For ablation studies and baseline comparisons, we focus on evaluating the performance of the privileged policy, as the deployable student policy is trained solely through imitation of the privileged teacher. For real-world experiments, we deploy our policy on Unitree G1~\citep{unitree_g1}, a medium-sized humanoid robot with $23$ DoFs and a height of $1.32$ meters.
The policy tracking performance is quantitatively evaluated with: \textbf{1)} $E_{\text{mpkpe}}$, mean per keybody position error, in $mm$; \textbf{2)} $E_{\text{mpjpe}}$, mean per joint position error, in $rad$; \textbf{3)} $E_{\text{vel}}$, linear velocity error, in $m/s$; \textbf{4)} $E_{\text{yaw vel}}$, yaw velocity error, in $rad/s$.

\subsection{Baselines}
We compare \ours's performance with ExBody2~\citep{ji2024exbody2} in simulation. We re-implement ExBody2 and train it on our filtered dataset. From \tabref{tab:ablations-baselines}, we found that \ours outperforms ExBody2 on both local tracking performance ($E_{\text{mpkpe}}$ and $E_{\text{mpjpe}}$) and global tracking performance ($E_{\text{vel}}$ and $E_{\text{yaw vel}}$).

\input{tables/ablation_baseline}

\subsection{Ablation Studies}
For this part, we conduct ablation studies to investigate the contribution of each component. 
Specifically, to evaluate the effects of the Motion MoE architecture and Adaptive Sampling strategy, we consider ablations include: \textbf{1) \ours w.o. A.S. \& MoE:} MoE model is replaced with an MLP of equivalent number of parameters and Adaptive Sampling is removed. \textbf{2) \ours w.o. MoE:} Only MoE model is replaced with a MLP of the same size. \textbf{3) \ours w.o. A.S.:} Only Adaptive Sampling is removed during training.

Additionally, we investigate the impact of motion input configuration with ablations: \textbf{\ours-M:} Only the immediate next frame of motion is provided as input to the policy network; \textbf{\ours-Lx-M:} Both the immediate next frame and a window of \textit{x} seconds of future motion frames are input to the policy network; \textbf{\ours-Lx:} Only a window of \textit{x} seconds of future motion frames is provided, without including the immediate next frame.

\begin{figure}[!t]
     \centering
     \includegraphics[width=0.98\textwidth]{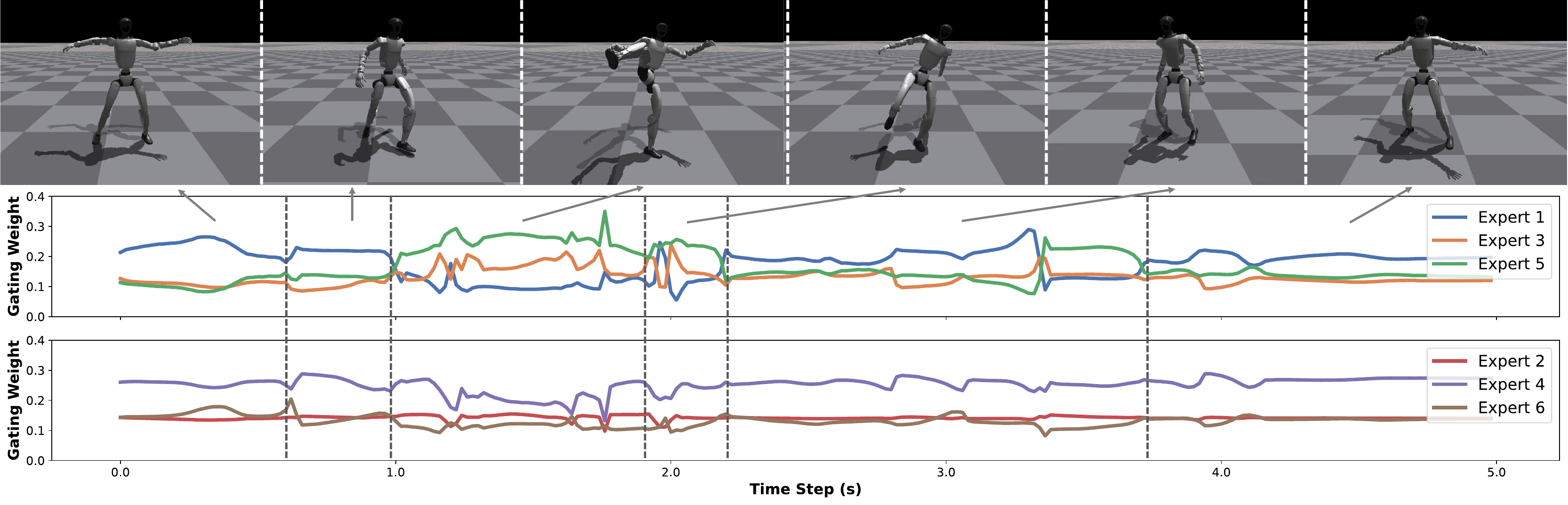}
    \caption{Plot of the output of gating network with respect to time on a motion clip composed of a sequence of skills.}
    \label{fig:moe-vis}
\end{figure}

\subsubsection{Motion MoE}
\begin{figure}[!t]
    \centering
    \includegraphics[width=0.99\textwidth]{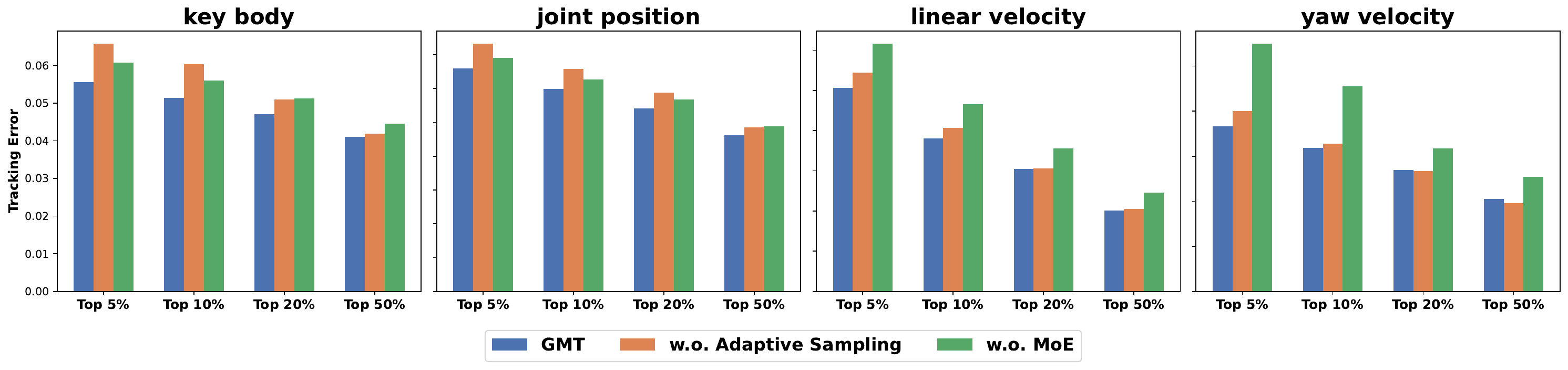}
    \caption{Top percentile tracking errors on the whole AMASS dataset.}
    \label{fig:amass-error}
\end{figure}

As shown in \tabref{tab:ablations-baselines}(a), \ours outperforms the baselines in tracking performance on both the AMASS test set and the entire LAFAN1 dataset. Furthermore, as shown in \Figref{fig:amass-error}, where top percentile tracking errors on AMASS are recorded. Both the statistics and the figure indicate that MoE helps improve a lot more on more challenging motions. For qualitative evaluation, \Figref{fig:moe-vis}, visualized the expert selection on a composite motion sequence consisting of standing, kicking, walking backward for a few steps, and standing again. The gating weights of each expert over time show clear transitions in expert activation across different phases of the motion. This suggests that individual experts specialize in different types of motion, validating the intended role of the MoE structure in capturing motion diversity.

\subsubsection{Adaptive Sampling}
\begin{figure}[!ht]
    \centering
    \begin{minipage}[b]{0.51\textwidth}
        \centering
        \includegraphics[width=\linewidth]{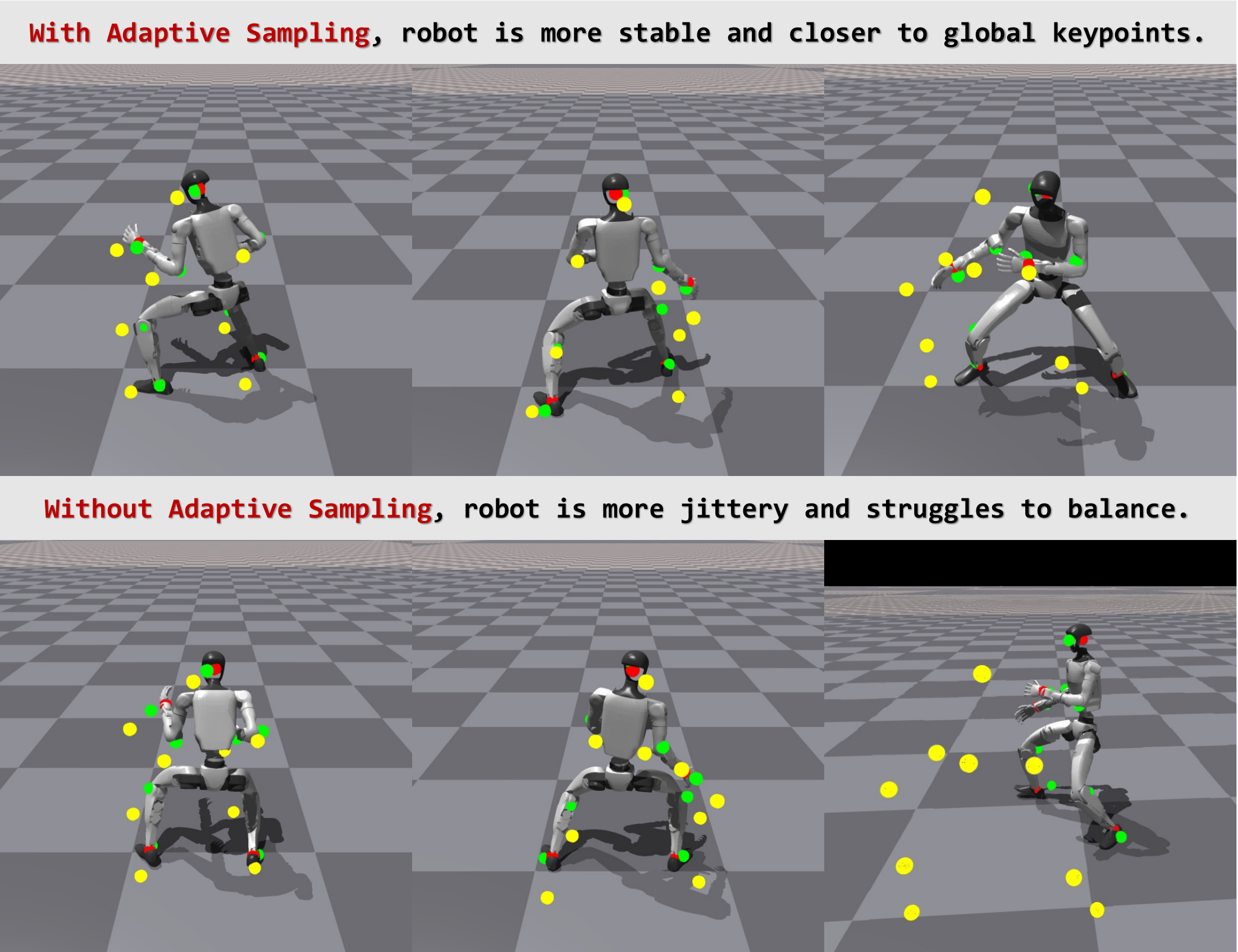}
        \vspace{2mm}
        {\small \textbf{(a)}}
        \label{fig:vis-sampling-comp}
    \end{minipage}
    \hfill
    \begin{minipage}[b]{0.46\textwidth}
        \centering
        \includegraphics[width=\linewidth]{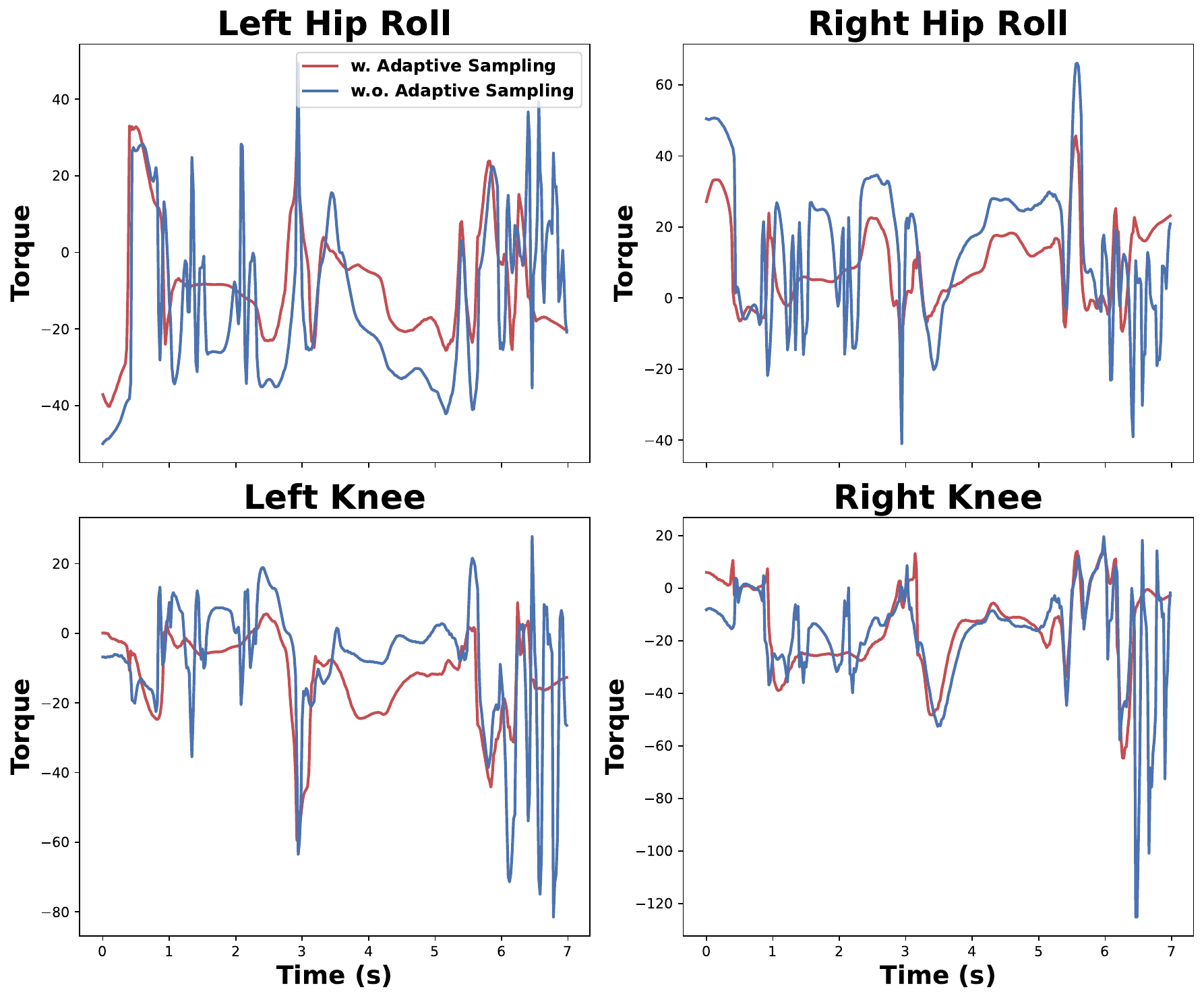}
        \vspace{2mm}
        {\small \textbf{(b)}}
        \label{fig:plot-sampling-comp}
    \end{minipage}
    \caption{The performance of policies with and without Adaptive Sampling on one segment extracted from a long motion clip. (a) Visualization of policy performance in the simulator. (b) Torque outputs of several joints corresponding to this segment.}
    \label{fig:comp-adaptive-sampling}
\end{figure}

As shown in \tabref{tab:ablations-baselines}(a), Adaptive Sampling effectively improved the tracking performance on both datasets. Similar to Motion MoE, as shown in \Figref{fig:amass-error}, Adaptive Sampling improves more on more challenging motions. To qualitatively evaluate the influence of this strategy,
 we extract a short clip from a long and composite $240$s motion, and compare the performance of policies trained with and without Adaptive Sampling in \Figref{fig:comp-adaptive-sampling}(a). In addition, we plot the torque of key joints, including the knee and hip roll, in \Figref{fig:comp-adaptive-sampling}(b). The results show that without Adaptive Sampling, the policy fails to learn this clip with high-quality and struggles to balance. These artifacts make real-world deployment impossible. 

\subsubsection{Motion Inputs}
Experimental results in \tabref{tab:ablations-baselines}(b) show that increasing the motion input window length leads to improved tracking accuracy. However, the performance of \textbf{\ours-L2} shows a significant degradation, indicating that inputting the immediate next frame into the policy network is crucial as well. This can be explained as that while a sequence of future frames captures the overall tendency of upcoming motions, it can lose some detailed information. In this way, inputting the immediate next frame into the policy can greatly enhance tracking performance by providing the nearest relevant information.

\subsection{Real-World Deployment}
As shown in \Figref{fig:teaser}, we successfully deploy our policy on a real-world humanoid robot, reproducing a wide range of human motions — including stylized walking, high kicking, dancing, spinning, crouch walking, soccer kicking, and many others — with high fidelity and state-of-the-art performance. For more detailed deployment results, please refer to the \href{https://gmt-humanoid.github.io}{project website}.

\subsection{Applications - Tracking MDM-Generated Motions}
\begin{figure}[!t]
    \centering
    \includegraphics[width=0.98\textwidth]{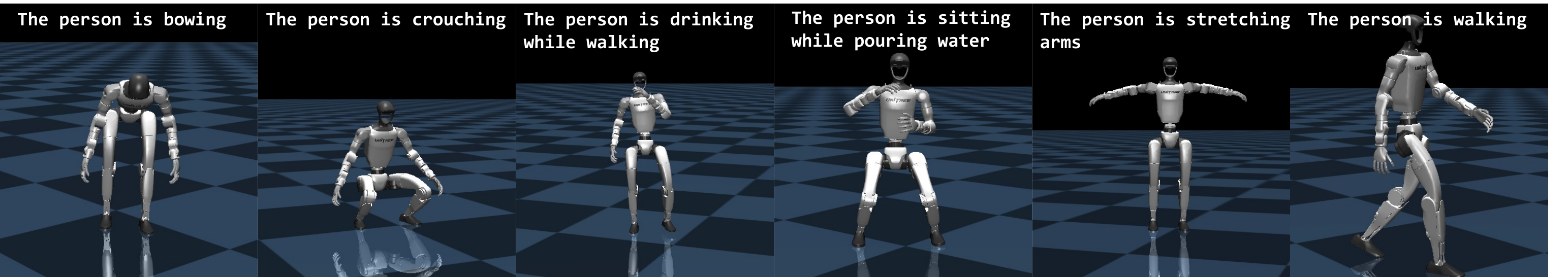}
    \caption{Motion tracking on MDM-generated motions.}
    \label{fig:mdm-track}
    \vspace{-0.5cm}
\end{figure}

We test our policy with motion diffusion model (MDM)~\citep{tevet2022human} generated motions in MuJoCo~\citep{todorov2012mujoco} sim-to-sim settings. Results in \Figref{fig:mdm-track} show that \ours can perform well on motions generated with MDM by text prompts, proving the potential of \ours to be applied to other downstream tasks.

\section{Conclusion}
\label{sec:conclusion}
In this paper, we introduce \ours, a general and scalable motion tracking framework that trains a single unified policy to enable humanoid robots to imitate diverse motions in the real world. We conduct extensive experiments to evaluate the contribution of each component to the overall tracking performance and demonstrate that \ours can track motions from other resources like MDM. Real-world deployment results demonstrate that our general unified policy achieves state-of-the-art performance compared to prior works. We believe that this general controller can serve as a foundation for future whole-body algorithm development on humanoid robots.

\clearpage
\section{Limitations}
While \ours achieves state-of-the-art performance as a unified, general motion tracking controller, it still has several limitations:
\begin{itemize}
    \item \textbf{Lack of Contact-Rich Skills.} Due to the significant additional simulation complexity required to model contact-rich behaviors~\citep{he2025getup}, along with hardware limitations, our framework does not currently support skills such as getting up from a fallen state or rolling on the ground.
    \item \textbf{Limitations on Challenging Terrains.} Our current policy is trained without any terrain observations and is not designed for imitation on challenging terrains like slopes and stairs. In future work, we aim to extend our framework to develop a general and robust controller capable of operating across both flat and challenging terrains.
\end{itemize}




\bibliography{ref}  

\newpage
\input{appendix}

\end{document}

%% file: math_commands.tex
\usepackage{amsmath,amsfonts,bm}
\usepackage{amsfonts}

\def\tabref#1{table~\ref{#1}}

\def\Figref#1{Fig.~\ref{#1}}

\def\eqref#1{equation~\ref{#1}}

\def\1{\bm{1}}

\def\va{{\bm{a}}}

\def\ve{{\bm{e}}}

\def\vg{{\bm{g}}}

\def\vo{{\bm{o}}}
\def\vp{{\bm{p}}}
\def\vq{{\bm{q}}}
\def\vr{{\bm{r}}}
\def\vs{{\bm{s}}}

\def\vv{{\bm{v}}}

\def\vz{{\bm{z}}}

\DeclareMathAlphabet{\mathsfit}{\encodingdefault}{\sfdefault}{m}{sl}
\SetMathAlphabet{\mathsfit}{bold}{\encodingdefault}{\sfdefault}{bx}{n}

\newcommand{\R}{\mathbb{R}}



%% file: tables/work_comp.tex
\begin{table}[!ht]
\caption{Comparison of closely related works.}
\label{tab:work-comp}
\centering
\rowcolors{2}{white}{gray!15}
\begin{tabular}{ccccc}
\toprule
\textbf{Method} & \textbf{Single Policy} & \textbf{Whole-Body} & \textbf{Diverse} & \textbf{Dataset Size} \\
\midrule
ExBody~\citep{cheng2024expressive} & \cmark & \xmark & \xmark & $780$ clips  \\
ExBody2~\citep{ji2024exbody2} & \cmark & \cmark & \cmark & $1050$ clips(filtered) \\
HumanPlus~\citep{fu2024humanplus} & \cmark & \cmark & \xmark & $10$k clips \\
OmniH2O~\citep{he2024omnih2o} & \cmark & \cmark &  \xmark & $14$k clips (augmented) \\
ASAP~\citep{he2025asap} & \xmark & \cmark & \cmark & -- (internet clips) \\
\textbf{\ours(Ours)} & \cmark & \cmark & \cmark & $8925$ clips (filtered) \\ 
\bottomrule
\end{tabular}
\end{table}

%% file: tables/ablation_baseline.tex
\begin{table}[!t]
\caption{Simulation evaluation of trained policies on AMASS-test and LAFAN dataset. All policies are trained on our filtered AMASS+LAFAN dataset. For all the baseline comparison and ablation studies, we only compare the performance of privileged policies.}
\label{tab:ablations-baselines}
\centering
\renewcommand{\arraystretch}{0.9}
\resizebox{1.0\textwidth}{!}{
\begin{tabular}{lcccccccc}
\toprule
\multicolumn{1}{l}{} & \multicolumn{4}{c}{\textbf{AMASS-Test}} & \multicolumn{4}{c}{\textbf{LAFAN1}} \\
\cmidrule(r){1-1}\cmidrule(r){2-5}\cmidrule(r){6-9}
\textbf{Method} & $E_{\text{mpkpe}}$ $\downarrow$ & $E_{\text{mpjpe}}$ $\downarrow$ & $E_{\text{vel}}$ $\downarrow$ & $E_{\text{yaw vel}}$ $\downarrow$ & $E_{\text{mpkpe}}$ $\downarrow$ & $E_{\text{mpjpe}}$ $\downarrow$ & $E_{\text{vel}}$ $\downarrow$ & $E_{\text{yaw vel}}$ $\downarrow$ \\
\cmidrule(r){1-1}\cmidrule(r){2-5}\cmidrule(r){6-9}
\rowcolor{lightgray}
\multicolumn{1}{l}{\textbf{Teacher \& Student}} &&&&&&&&\\
\cmidrule(r){1-1}\cmidrule(r){2-5}\cmidrule(r){6-9}
Privileged Policy &$42.07$&$0.0834$&$0.1747$&$0.2238$&$45.16$&$0.0975$&$0.2837$&$0.3314$ \\
Student Policy &$42.01$&$0.0807$&$0.1943$&$0.2211$&$46.14$&$0.1060$&$0.3009$&$0.3489$ \\
\cmidrule(r){1-1}\cmidrule(r){2-5}\cmidrule(r){6-9}
\rowcolor{lightgray}
\multicolumn{1}{l}{\textbf{Baseline}} &&&&&&&&\\
\cmidrule(r){1-1}\cmidrule(r){2-5}\cmidrule(r){6-9}
ExBody2~\citep{ji2024exbody2} &
$50.28{\scriptstyle\pm0.28}$&$0.0925{\scriptstyle\pm0.001}$&$0.1875{\scriptstyle\pm0.001}$&$0.3402{\scriptstyle\pm0.004}$&$58.36{\scriptstyle\pm0.48}$&$0.1378{\scriptstyle\pm0.002}$&$0.3461{\scriptstyle\pm0.005}$&$0.4260{\scriptstyle\pm0.006}$
 \\
\ours(ours) &$\mathbf{42.07 {\scriptstyle\pm0.17}}$&$\mathbf{0.0834{\scriptstyle\pm0.001}}$&$\mathbf{0.1747{\scriptstyle\pm0.001}}$&$\mathbf{0.2238{\scriptstyle\pm0.002}}$&$\mathbf{45.16{\scriptstyle\pm0.35}}$&$\mathbf{0.0975{\scriptstyle\pm0.001}}$&$\mathbf{0.2837{\scriptstyle\pm0.004}}$&$\mathbf{0.3314{\scriptstyle\pm0.003}}$ \\
\cmidrule(r){1-1}\cmidrule(r){2-5}\cmidrule(r){6-9}
\rowcolor{lightgray}
\multicolumn{1}{l}{\textbf{(a) Ablations}} &&&&&&&&\\
\cmidrule(r){1-1}\cmidrule(r){2-5}\cmidrule(r){6-9}
\ours w.o. MoE &$42.53{\scriptstyle\pm0.19}$&$0.0874{\scriptstyle\pm0.00}$&$0.1902{\scriptstyle\pm0.002}$&$0.2483{\scriptstyle\pm0.001}$&$48.26{\scriptstyle\pm0.29}$&$0.1019{\scriptstyle\pm0.001}$&$0.3111{\scriptstyle\pm0.003}$&$0.3795{\scriptstyle\pm0.005}$ \\
\ours w.o. A.S. &$43.54{\scriptstyle\pm0.23}$&$0.0872{\scriptstyle\pm0.001}$&$0.2064{\scriptstyle\pm0.001}$&$0.2593{\scriptstyle\pm0.001}$&$49.61{\scriptstyle\pm0.30}$&$0.1041{\scriptstyle\pm0.002}$&$0.3019{\scriptstyle\pm0.003}$&$0.3574{\scriptstyle\pm0.003}$ \\
\ours w.o. A.S. \& MoE &$44.34{\scriptstyle\pm0.21}$&$0.0920{\scriptstyle\pm0.001}$&$0.2121{\scriptstyle\pm0.001}$&$0.2534{\scriptstyle\pm0.001}$&$52.34{\scriptstyle\pm0.33}$&$0.1110{\scriptstyle\pm0.002}$&$0.3263{\scriptstyle\pm0.003}$&$0.3584{\scriptstyle\pm0.007}$ \\
\ours (ours) &$\mathbf{42.07 {\scriptstyle\pm0.17}}$&$\mathbf{0.0834{\scriptstyle\pm0.001}}$&$\mathbf{0.1747{\scriptstyle\pm0.001}}$&$\mathbf{0.2238{\scriptstyle\pm0.002}}$&$\mathbf{45.16{\scriptstyle\pm0.35}}$&$\mathbf{0.0975{\scriptstyle\pm0.001}}$&$\mathbf{0.2837{\scriptstyle\pm0.004}}$&$\mathbf{0.3314{\scriptstyle\pm0.003}}$ \\
\cmidrule(r){1-1}\cmidrule(r){2-5}\cmidrule(r){6-9}
\rowcolor{lightgray}
\multicolumn{1}{l}{\textbf{(b) Motion Inputs}} &&&&&&&&\\
\cmidrule(r){1-1}\cmidrule(r){2-5}\cmidrule(r){6-9}
\ours-M &$46.02{\scriptstyle\pm0.25}$&$0.0942{\scriptstyle\pm0.001}$&$0.2282{\scriptstyle\pm0.001}$&$0.3311{\scriptstyle\pm0.003}$&$51.16{\scriptstyle\pm0.34}$&$0.1069{\scriptstyle\pm0.002}$&$0.3476{\scriptstyle\pm0.001}$&$0.4890{\scriptstyle\pm0.007}$ \\
\ours-L$0.5$-M &$43.64{\scriptstyle\pm0.19}$&$0.0855{\scriptstyle\pm0.001}$&$0.2051{\scriptstyle\pm0.001}$&$0.2439{\scriptstyle\pm0.001}$&$49.87{\scriptstyle\pm0.32}$&$0.1032{\scriptstyle\pm0.002}$&$0.3346{\scriptstyle\pm0.005}$&$0.3648{\scriptstyle\pm0.003}$ \\
\ours-L$1$-M &$43.15{\scriptstyle\pm0.22}$&$0.0867{\scriptstyle\pm0.001}$&$0.1989{\scriptstyle\pm0.002}$&$0.2465{\scriptstyle\pm0.001}$&$47.41{\scriptstyle\pm0.35}$&$0.1007{\scriptstyle\pm0.002}$&$0.3047{\scriptstyle\pm0.003}$&$0.3513{\scriptstyle\pm0.002}$ \\
\ours-L$2$ &$49.52{\scriptstyle\pm0.27}$&$0.1016{\scriptstyle\pm0.002}$&$0.2201{\scriptstyle\pm0.001}$&$0.2888{\scriptstyle\pm0.003}$&$61.24{\scriptstyle\pm0.42}$&$0.1368{\scriptstyle\pm0.002}$&$0.3925{\scriptstyle\pm0.008}$&$0.5558{\scriptstyle\pm0.009}$ \\
\ours-L$2$-M (ours) &$\mathbf{42.07 {\scriptstyle\pm0.17}}$&$\mathbf{0.0834{\scriptstyle\pm0.001}}$&$\mathbf{0.1747{\scriptstyle\pm0.001}}$&$\mathbf{0.2238{\scriptstyle\pm0.002}}$&$\mathbf{45.16{\scriptstyle\pm0.35}}$&$\mathbf{0.0975{\scriptstyle\pm0.001}}$&$\mathbf{0.2837{\scriptstyle\pm0.004}}$&$\mathbf{0.3314{\scriptstyle\pm0.003}}$ \\
\bottomrule 
\end{tabular}}
\vspace{-0.35cm}
\end{table}

%% file: appendix.tex
\section*{Appendix}

\subsection{Goal-Conditioned Reinforcement Learning}
In this work, motion tracking problem is defined as a goal-conditioned RL problem where given the goal, the agent interacts with the environment according to the policy $\pi$ to maximize an objective function~\citep{sutton1998reinforcement}. At each timestep $t$, the agent's policy takes the state $\vs_t$ and the goal $\vg_t$ as input, and outputs the action $\va_t$, which is formulated as $\pi(\va_t|\vs_t, \vg_t)$. When applied to the environment, the action $\va_t$ leads to the next state $\vs_{t+1}$ according to the environment dynamics $p(\vs_{t+1}|\vs_t, \va_t)$. A reward $r(\vs_{t+1}, \vs_{t}, \va_{t})$ is received at each timestep. The agent's goal is to maximize the expected return:
\begin{equation}
J(\pi)=\mathbb{E}_{p(\tau \mid \pi)}\left[\sum_{t=0}^{T-1} \gamma^t r_t\right],
\label{eq:RL-obj}
\end{equation}
where $p(\tau|\pi)$ represents the likelihood of the trajectory $\tau$, $T$ denotes the time horizon, and $\gamma$ is the discount factor.

\subsection{Sim-to-Real Transfer}
To enable successful sim-to-real transfer, we apply domain randomizations during the training of both teacher and student policies~\citep{peng2018dynamicsrand, rudin2022learning}. To further align simulation and real-world physical dynamics,we explicitly model the effect of the reduction drive's moment of inertia. Specifically, given a reduction ratio $k$, and a reduction drive moment of inertia $I$, we configure the \texttt{armature} parameter in the simulator as:
\begin{equation}
    \label{eq:sim2real}
    \texttt{armature} = k^{2}I,
\end{equation}
to approximate the effective inertia introduced by the reduction drive.

\subsection{Policy Learning}
Following previous works~\citep{ji2024exbody2, he2024omnih2o}, we adopt a two-stage training framework. In the first stage, we train a privileged teacher policy that observes both proprioceptive and privileged information, and outputs joint target actions $\hat{\va}_t$, optimized using PPO~\citep{schulman2017proximal}. In the second stage, we train a deployable student policy that takes as input a sequence of proprioceptive observation history, and is supervised by the teacher policy through DAgger~\citep{ross2011dagger}. The student policy is optimized by minimizing the $\ell_2$ loss between its output $\va_t$ and the teacher's output $\hat{\va_t}$ according to $\| \hat{\va_t} -\va \|_2^2$.

\subsection{Observations and Actions}
For the teacher policy observations consist of proprioception $\vo_t$, privileged information $\ve_t$, and motion targets $\vg$. $\vo_t$ consists of root angular velocity (3 dims), root roll and pitch (2 dims), joint positions (23 dims), joint velocities (23 dims), and last action (23 dims). $\ve_t$ consists of root linear velocity (3 dims), root height (1 dim), key body positions, feet contact mask (2 dims), mass randomization params (6 dim), and motor strength (46 dims). For the student policy, observation consists of proprioception $\vo_t$, proprioception history $\vo_{t-20}, \dots, \vo_{t}$, and motion targets $\vg$.

The output action is target joint positions.

\subsection{Training Details}
We use IsaacGym~\citep{makoviychuk2021isaacgym} as the physics simulator and the number of parallel environments is $4096$. We train the privileged policy for around $3$ days on an RTX4090 GPU, and then train the student policy for around $1$ day. The simulation frequency is $500$Hz and the control frequency is $50$Hz. The trained policy is validated in Mujoco~\citep{todorov2012mujoco} before deploying onto the real robot.

\subsection{Reward Functions}
Reward functions of first-stage training. Here $\vq$ denotes joint positions, $\dot\vq$ denotes joint veocities, $\ddot{\vq}$ denotes joint accelerations, $\vr$ denotes root rotations, $\vv$ denotes root velocites, $h$ denotes root height, and $\vp$ denotes key body positions. Full definitions are in \tabref{tab:rew-functions}.

\begin{table}[!ht]
    \centering
    \caption{Definitions of Reward Functions.}
    \begin{tabular}{cc}
    \toprule
    \textbf{Name} & \textbf{Definitions} \\ 
    \midrule
    tracking joint positions & $\exp(-\|\vq^{\text{ref}}_{t} - \vq_{t} \|_2^2)$  \\
    tracking joint velocities & $\exp(-\|\dot{\vq^{\text{ref}}_{t}} - \dot{\vq}_{t} \|_2^2)$ \\
    tracking root pose & $\exp(-\|\vr^{\text{ref}}_t - \vr_{t} \|_2^2 - \| h^{\text{ref}}_t - h_t \|_2^2)$ \\
    tracking root vel & $\exp(-\|\vv^{\text{ref}}_{t} - \vv_{t} \|_2^2)$ \\
    tracking key body positions & $\exp(-\|\vp^{\text{ref}}_{t} - \vp_{t} \|_2^2)$ \\
    alive & $1.0$ \\
    foot slip & $-\|\vv^{\text{foot}}_{t}F_{t}^{\text{foot contact}}\|_2$ \\
    joint velocities & $-\|\dot{\vq}_{t}\|$ \\
    joint accelerations & $-\|\ddot{\vq}_{t}\|$ \\
    action rate & $-\| \va_{t} - \va_{t-1} \|$ \\
    \bottomrule
    \end{tabular}
    \label{tab:rew-functions}
\end{table}

\subsection{Domain Randomizations}
To tackle sim-to-real gap, we apply extensive domain randomizations during training time. The details are shown in \tabref{tab:details-dr}.

\begin{table}[!ht]
    \centering
    \caption{Details of domain randomizations.}
    \begin{tabular}{cc}
    \toprule
    \textbf{Name} & \textbf{Range} \\ 
    \midrule
    Terrain Height & $[0, 0.02]m$  \\
    Gravity & $[-0.1, 0.1]$ \\
    Friction & $[0.1, 2.0]$ \\
    Robot Base Mass & $[-3, 3]kg$ \\
    Robot Base Mass Center & $[-0.05, 0.05]m$ \\
    Push Velocity & $[0.0, 1.0]m/s$\\
    Motor Strength & $[0.8, 1.2]$ \\
    Action Delay & $[0, 0.02]s$ \\
    \bottomrule
    \end{tabular}
    \label{tab:details-dr}
\end{table}
